\pdfoutput=1

\documentclass[11pt]{article}

\usepackage{EACL2023}

\usepackage{times}
\usepackage{latexsym}
\usepackage{graphicx}
\usepackage{amsmath,amssymb,amsfonts}
\usepackage{mathtools}
\usepackage{arydshln}

\usepackage[T1]{fontenc}

\usepackage[utf8]{inputenc}

\usepackage{microtype}

\usepackage{inconsolata}

\title{Towards Fine-tuning Pre-trained Language Models with\\Integer Forward and Backward Propagation}

\author{Mohammadreza Tayaranian$^{*1}$ \quad Alireza Ghaffari$^{*1}$ \quad Marzieh S. Tahaei$^1$ \\
  {\bf Mehdi Rezagholizadeh$^1$} \quad {\bf Masoud Asgharian$^2$} \quad {\bf Vahid Partovi Nia$^1$} \\
  $^1$ Huawei Noah’s Ark Lab, Montreal Research Center \\
  $^2$ Department of Mathematics and Statistics, McGill University \\
  \texttt{\{mohammadreza.tayaranian, alireza.ghaffari, marzieh.tahaei\}@huawei.com} \\
  \texttt{\{mehdi.rezagholizadeh, vahid.partovinia\}@huawei.com}  \\
  \texttt{masoud.asgharian2@mcgill.ca} \\ }

\begin{document}
\maketitle
\begin{abstract}
The large number of parameters of some prominent language models, such as BERT, makes their fine-tuning on downstream tasks computationally intensive and energy hungry.
Previously researchers were focused on lower bit-width integer data types for the forward propagation of language models to save memory and computation.
As for the backward propagation, however, only 16-bit floating-point data type has been used for the fine-tuning of BERT.
In this work, we use integer arithmetic for both forward and back propagation in the fine-tuning of BERT.
We study the effects of varying the integer bit-width on the model's metric performance.
Our integer fine-tuning uses integer arithmetic to perform forward propagation and gradient computation of linear, layer-norm, and embedding layers of BERT.
We fine-tune BERT using our integer training method on SQuAD v1.1 and SQuAD v2., and GLUE benchmark.
We demonstrate that metric performance of fine-tuning 16-bit integer BERT matches both 16-bit and 32-bit floating-point baselines.
Furthermore, using the faster and more memory efficient 8-bit integer data type, integer fine-tuning of BERT loses an average of 3.1 points compared to the FP32 baseline.


\end{abstract}

\section{Introduction}

\def\thefootnote{*}\footnotetext{Equal contribution.}\def\thefootnote{\arabic{footnote}}

Over the past few years, integration of attention mechanisms into deep learning models led to the creation of transformer based models.
BERT \cite{devlin2018bert} is a prominent transformer based language model which has shown state-of-the-art performance in natural language processing (NLP) tasks.

BERT requires high memory and computational resources due to its large number of parameters.
Having large number of parameters incurs challenges for inference, training, and also fine-tuning of this model.
Moreover, the training phase i.e. pre-training and fine-tuning, involves more operations compared to the inference.
More specifically, the training phase includes gradient computation and weight update that make the training more computationally intensive.

\begin{figure}[t]
\centering
\includegraphics[width=0.9\columnwidth]{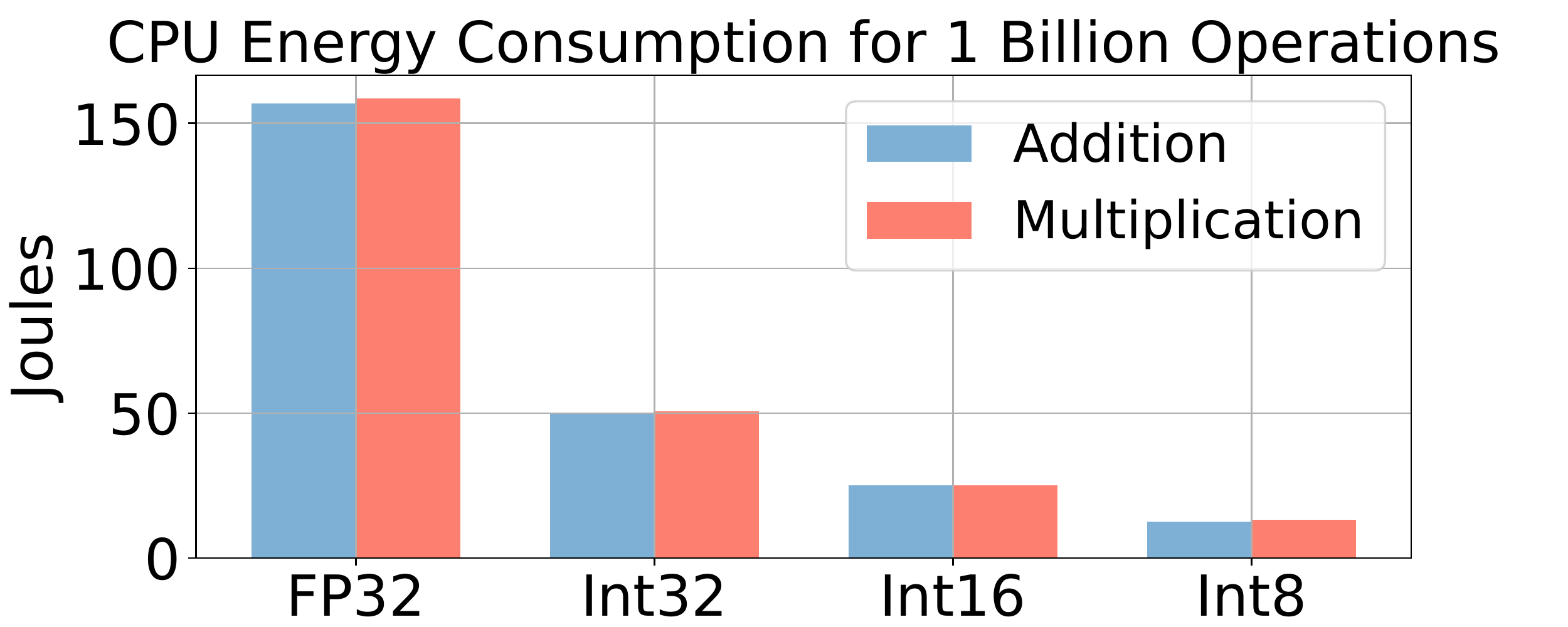}
\includegraphics[width=0.9\columnwidth]{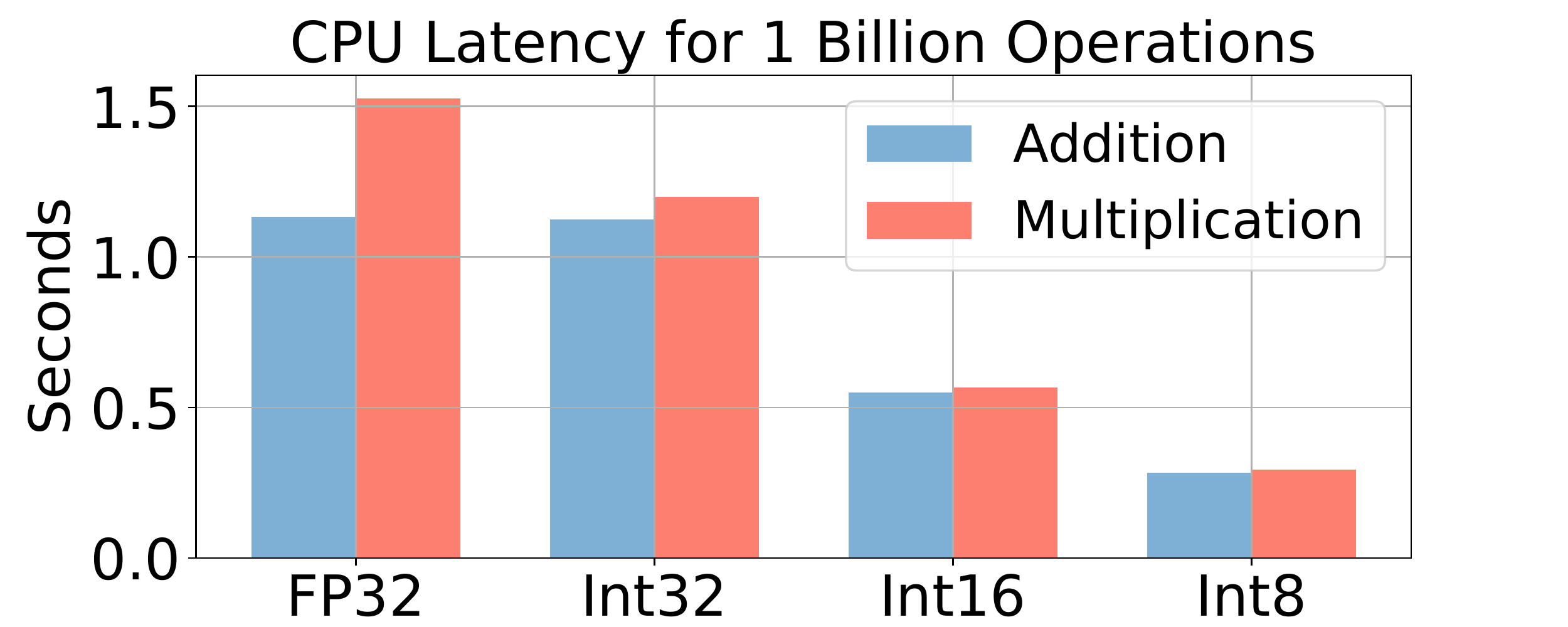}
\caption{Energy consumption and latency of 1 billion operations using various data types, measured on an Intel\textsuperscript{\textregistered} Xeon\textsuperscript{\textregistered}  CPU E5-2698 v4.}
\label{fig:energy}
\end{figure}

One method of reducing the computational complexity of deep learning models is to represent their parameters and activations in low bit-width data types.
This reduces the memory footprint of the model and enables more efficient computations. 
For instance, Figure \ref{fig:energy} shows that low-bit integer data types have higher throughput and better energy consumption compared to floating-point.

Previous research attempts at integer quantization of transformer based language models were only focused on forward propagation and the gradient computation were kept in 32-bit floating-point data type (FP32) \cite{bhandare2019efficient, kim2021bert, zafrir2019q8bert}.

Furthermore, earlier efforts for using low bit-width data types for gradient computation of transformer based language models has only been limited to 16-bit floating-point (FP16).
This method, known as mixed precision training \cite{micikevicius2017mixed}, uses FP16 data type to represent weights, activations and gradients while using FP32 for the weight update.

Here we present an integer fine-tuning method for transformer based language models such as BERT.
Unlike previous works, we use integer data types for both forward propagation and gradient computation during the fine-tuning of BERT.
Moreover, we use the dynamic fixed-point format to represent floating-point numbers as integers.

Our integer mapping strategy can be used alongside floating-point numbers in fine-tuning and inference similar to mixed precision training.
In our proposed strategy, the arithmetic of all the compute intensive layers for both forward and back propagation are performed using integer arithmetic while other components of the model, such as nonlinear functions and the weight updates are kept in FP32.
We use integer versions of compute intensive layers such as linear, normalization (layer-norm), and embedding layers.

We study the effect of various bit-widths of the integer input activation and show that increasing the bit-width of the fixed-point mapping function improves the convergence behaviour of the model. This enables us to find the minimum bit-width required for integer fine-tuning of BERT.

Our fine-tuning experiments show that 16-bit integer BERT is able to match the metric performance of mixed precision FP16 and FP32 methods.

We also further reduce the bit-widths and show that integer fine-tuning of BERT with 8-bit integer weights and 12-bit integer activations has a score drop of $3.1$ compared to the original performance.

To summarize, this paper makes the following contributions:

\begin{itemize}
    
    \item Integer fine-tuning of transformer based language models that uses integer arithmetic for both the forward and back propagation of compute intensive layers such as linear, layer-norm, and embedding. To the best of our knowledge, this is the first time that integer data type is used for back propagation of pre-trained language models.
    
    \item Analyzing the effect of changing the bit-width of dynamic fixed-point format on the convergence of fine-tuning.~\textbf{Remark 3} discusses that the convergence behaviour of our integer fine-tuning is directly related to the variance of dynamic fixed-point mapping and is controlled by the bit-width.
    
    \item We show that fine-tuning BERT using 16-bit integer numbers is able to outperform the FP16 mixed precision fine-tuning method.
    
\end{itemize}

The rest of this paper is structured as follows.
Section \ref{sec:related} briefly discusses previous works in which low bit-width data types are used for inference and training of deep learning models.
Section \ref{sec:methodology} provides details of our integer fine-tuning method, including the representation mapping functions and integer-only layers.
The convergence behaviour of the dynamic fixed-point mapping is studied in Section \ref{sec:convergence} by providing empirical observations and theoretical analysis.
The fine-tuning experiments on various integer and floating-point setups are presented in Section \ref{sec:results}.
Finally, Section \ref{sec:conclusion} concludes the ideas proposed in this work.

\section{Related Works}
\label{sec:related}

In this section we discuss the previous works that use low bit-width data types in transformer based language models.
These works could be categorized into two major groups.
In the first group, called low-bit inference, the low bit-width data types are used only in the forward propagation phase to improve computational complexity and reduce memory usage during the inference.
In the second group, also known as low-bit training, lower bit-width data types are used for both the forward and back propagation phases.

\subsection{Low-bit Inference}

Previous research on low-bit inference quantize the model parameters and activations to speed up the forward propagation.
This category is itself divided into quantization-aware training (QAT) and post-training quantization (PTQ) methods.

In QAT, quantization is performed during training, allowing the model parameters to adapt to the quantization noise. QAT relies on high-precision FP32 gradients to train the model and adapt it to the quantization noise.

For instance, \cite{zafrir2019q8bert} proposed Q8BERT which quantizes the inference computations of all linear and embedding layers of BERT to 8-bit integers and updates the quantization scale with a moving average. Similarly, \cite{shen2020q} suggested Q-BERT which requires the computation of hessian matrix for each group of parameters to be used in a mixed precision fine-tuning with different bit-widths. \cite{kim2021bert} proposed I-BERT that uses a uniform quantization scheme to quantize input activations and weights of various components of BERT. In I-BERT, the quantization scaling factors are computed based on the distribution of the training data. 

Unlike QAT that performs quantization of inference operations during training, Post-Training Quantization (PTQ) methods apply quantization to the parameters when the training is completed. Thus, they require extra calibration or parameter tuning to adapt the model to the quantized parameters.

For instance, \cite{bhandare2019efficient} quantized the matrix multiplications of the original transformer architecture from \cite{vaswani2017attention} to 8-bit integer data type.
Moreover, the quantization is done only for the forward propagation and requires extra calibration using validation data to tune the boundaries of the quantization function.
\cite{zadeh2020gobo} introduced GOBO which compresses the fine-tuned weights of BERT by grouping them into two categories of Gaussian and outlier.
The outlier weights are kept in FP32, while the Gaussian weights are quantized to lower bits.
For lower bit-width regimes, TernaryBERT and BinaryBERT are able to push the quantization to 2 and 1 bits respectively \cite{zhang2020ternarybert, bai2020binarybert}.
They both rely on methods such as data augmentation and knowledge distillation to adapt the model to the low-bit weights.

\subsection{Low-bit Training}

Research on low-bit training try to perform both the forward propagation and gradient computation in low-bit arithmetic.
Using low precision number formats for gradients reduces the model's ability to adapt the parameters to the quantization noise, but increases the throughput and reduces the memory footprint.

FP16 mixed precision training \cite{micikevicius2017mixed} is a common method currently for low-bit fine-tuning of transformer based language models.
This method uses FP16 data type in both forward propagation and gradient computation, while using FP32 for the weight update.
Unlike FP16 mixed precision training, our work uses dynamic fixed-point format which allows for multiple choices of bit-width for the data type. We show that our 16-bit integer fine-tuning method outperforms FP16 mixed precision training in terms of metric score.

Using integer data types in the training of deep learning models has been previously studied for the computer vision tasks.
For instance, \cite{zhang2020fixed} quantized the input activations, gradients and parameters of the linear layers for various convolutional neural networks (CNN).
Similarly, \cite{zhao2021distribution} adapted the quantization parameters by detecting the distribution of the gradients in the channel dimension.
In both these works the quantization error is measured during training and is used to adjust the quantization scale, whereas our method does not require any information about  distribution of data or gradients.
\cite{zhu2020towards} applied a quantization scheme to train CNN architectures with ``direction sensitive gradient clipping'' and learning rate scaling to control the quantization error of gradients. Our integer fine-tuning method does not require gradient clipping and can follow the same loss trajectory as the floating-point baseline with the same hyper-parameters.
Our proposed method improves upon \cite{ghaffari2022integer} which uses dynamic fixed-point format for integer training of deep learning models.
Unlike \cite{ghaffari2022integer}, our work studies various bit-widths for both weights and activations to find the minimum bit-width required for fine-tuning BERT.
Furthermore, we study integer training method on large language models where low-bit quantization is known to be a challenging task \cite{bondarenko2021understanding}.
To the best of our knowledge, this is the first time where integer numbers are used for the back propagation of transformer based language models.


\begin{figure*}[t]
\centering
\includegraphics[width=1.8\columnwidth]{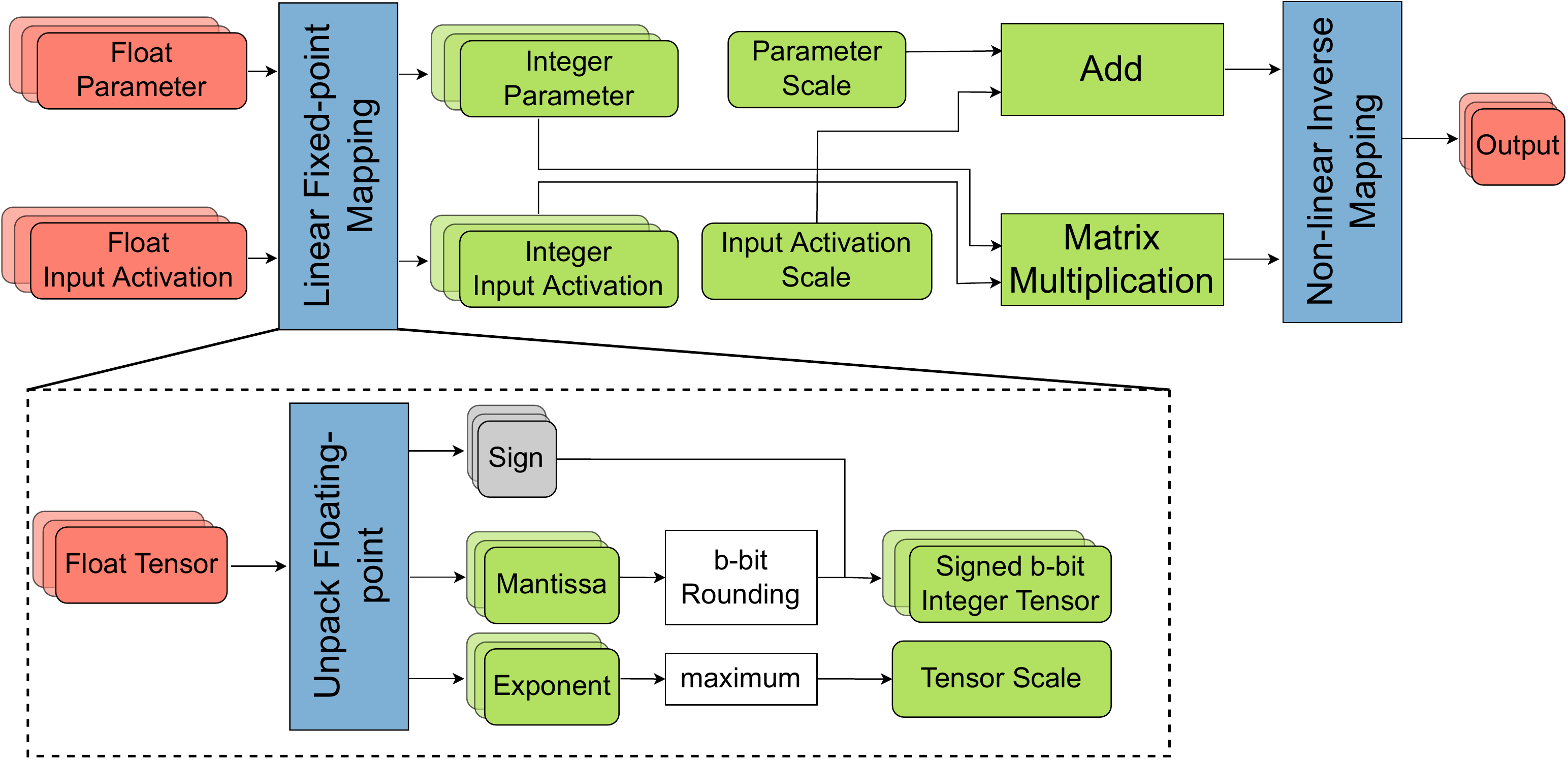} 
\caption{Forward propagation operations in an integer-only linear layer. Green boxes use integer arithmetic and red boxes use floating-point data type. Here, the integer output is generated using an integer matrix multiplication and the output scale is generated by a single add operation. The bottom panel shows the linear fixed-point mapping for the input tensors, that are the input activation and the parameter tensor in this figure.}
\label{fig:integer_linear}
\end{figure*}

\section{Methodology}
\label{sec:methodology}

\subsection{Representation Mapping}
\label{sec:mapping}

We use the dynamic fixed-point format \cite{williamson1991dynamically} to map the floating-point numbers to integer data type.
This format, also known as block floating-point, maps floating-point numbers to blocks of integer numbers, with each block having its unique scale.
For more information on various number formats refer to Appendix \ref{sec:appendix_datatypes}.

We use a linear fixed-point mapping function to map floating-point numbers to integer numbers.
The linear fixed-point mapping converts a floating-point tensor $\mathbf{F}$ to a tensor of integers and a single scale factor.

The integers are obtained by rounding the floating-point mantissas.
The scale is the maximum of the floating-point exponents of $\mathbf{F}$.
The bottom section of Figure \ref{fig:integer_linear} shows the internal operations of the linear fixed-point mapping.

To map the fixed-point numbers to floating-point, a non-linear inverse mapping function is used.
The inverse mapping converts integer numbers into normalized floating-point mantissas and packs each integer with its corresponding scale into a floating-point number.

Details of the representation mapping functions are provided in \cite{ghaffari2022integer}.
Our methodology differs in that it includes various bit-widths for both weights and activations for the fine-tuning of transformer based language models.
We exploit this mapping strategy to explore various bit-widths for weights and activations in order to find the minimum bit-width for fine-tuning the model.

\subsection{Integer Fine-tuning}

Our method uses integer arithmetic for weights, activations and gradients, while the weight update is kept in FP32.
Moreover, our proposed BERT setups use integer-only versions for all the linear, layer-norm and embedding layers in which internal operations are performed with integer arithmetic.

\subsubsection{Linear Layer}
Figure \ref{fig:integer_linear} depicts a high-level view of forward propagation operations of the integer-only linear layer.
All the parameters and activations of the layer are first mapped to dynamic fixed-point using the linear fixed-point mapping function.
In the case of linear layer, the integer parameters and input activations are then sent to an integer matrix multiplication function to generate the integer output.
If needed, the integer output could be mapped back to floating-point to be used by other layers of the model using the non-linear inverse mapping.

For back propagation, the gradients of the parameters and input activations are also computed using integer arithmetic.
Using integer matrix multiplication, the output gradients are multiplied by input activations and parameters to compute the gradients.
Since the weight update is performed in FP32, the integer gradients and their scales are passed to the non-linear inverse mapping to be mapped to FP32.

\subsubsection{Layer-norm}

The layer normalization or layer-norm performs the following operation on its input $X$ \cite{layernorm}:

\begin{equation}
    {\gamma}\frac{X - \mu}{\sqrt{\sigma^2 + \epsilon}} + {\beta}\text{.}\
    \label{eq:batchnorm}
\end{equation}

Here $\gamma$ and $\beta$ are the weight and bias parameters, and $\sigma$ and $\mu$ are input standard deviation and mean respectively.
For the forward propagation of integer layer-norm we map $X$ to dynamic fixed-point format and compute $\sigma$ and $\mu$ using integer arithmetic.
Note that multiplication to $\gamma$ and addition with $\beta$ are also performed using integer arithmetic.
Moreover, the back propagation also uses integer arithmetic to compute the gradients for the input, $\gamma$, and $\beta$.

\subsubsection{Embedding Layer}

The embedding layer is a lookup table that stores embeddings.
The layer takes a list of indices as input and returns the list of corresponding embeddings for each index.
The integer embedding layer, handles integer embeddings and needs less memory footprint to store these values.
For the back propagation,  the embedding layer applies the output integer gradients directly to each corresponding row of the lookup table.

\begin{figure}[t]
\centering
\includegraphics[width=0.9\columnwidth]{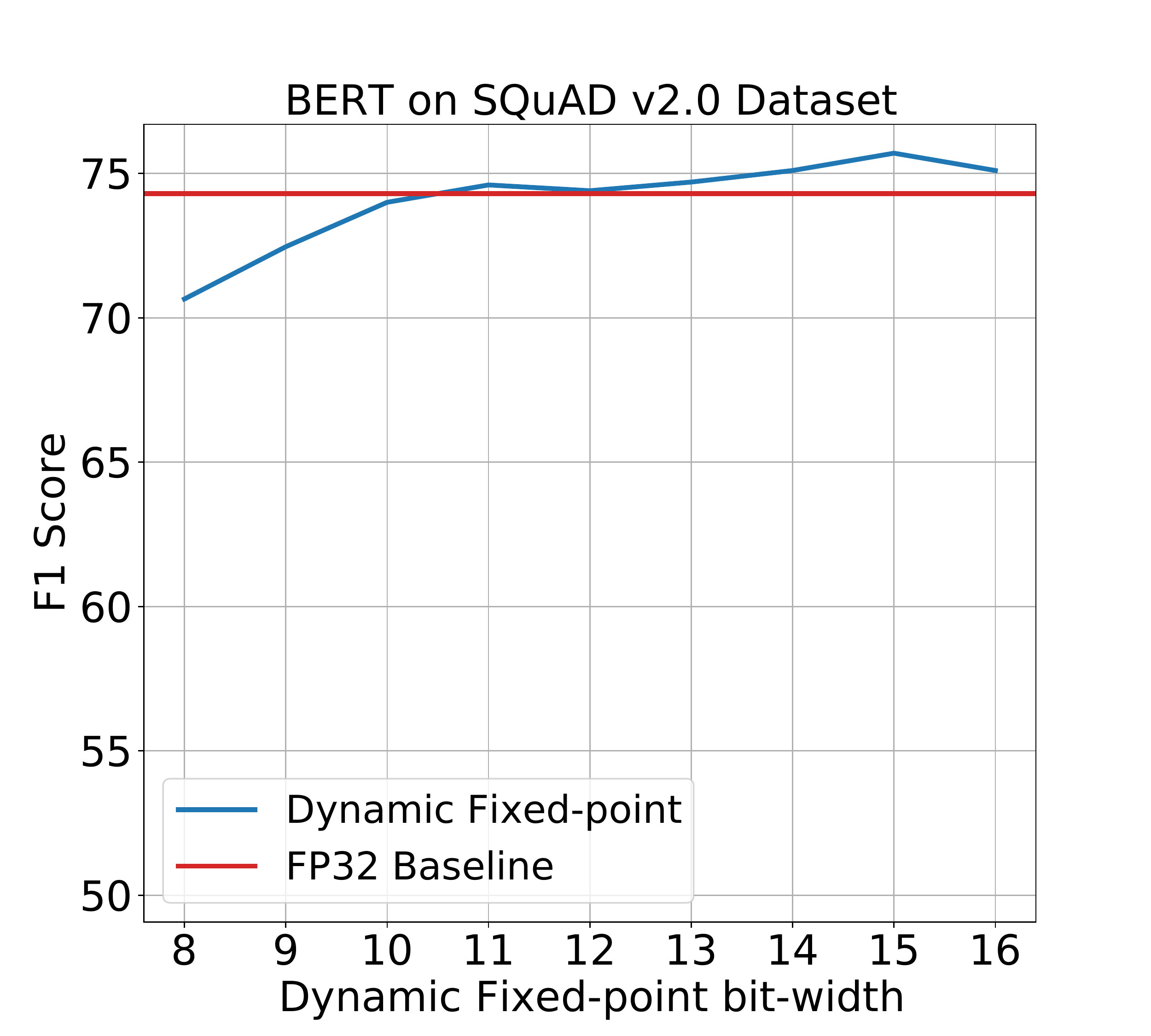}
\caption{F1 score of fine-tuning BERT using $b$-bit gradients, and activations on SQuAD v2.0 dataset. For the 8-bit and 9-bit fixed-point bit-widths, we use 12-bit input activations.}
\label{fig:ablation}
\end{figure}
\begin{figure}[t]
\centering
\includegraphics[width=0.9\columnwidth]{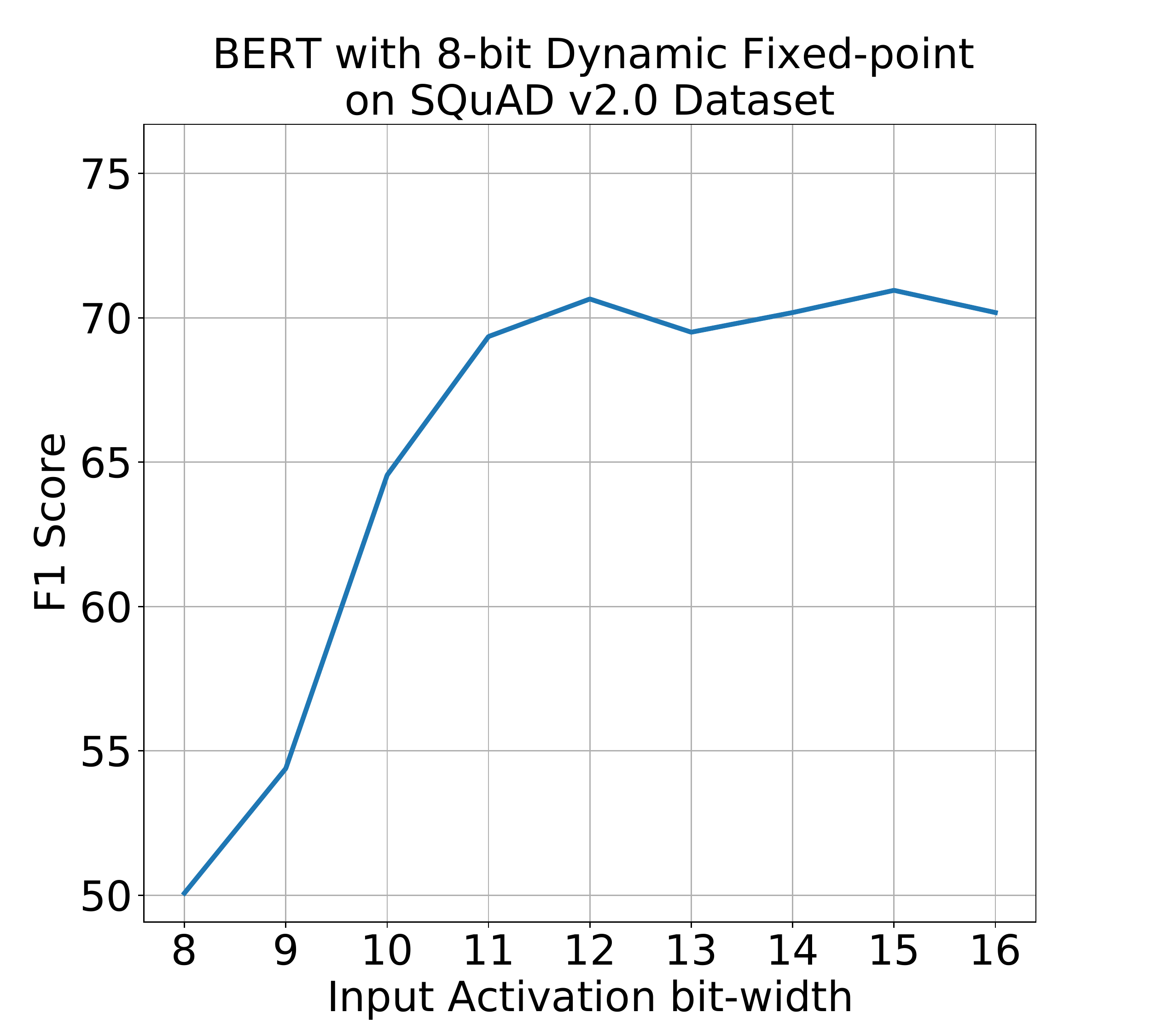}
\caption{F1 score of fine-tuning BERT using 8-bit weights and gradients,  with varying input activation bit-width on SQuAD v2.0 dataset. Note that \textbf{Remark 3} justifies this experiment  using the variance of $b$-bit dynamic fixed-point mapping. } 
\label{fig:ablation_activation}
\end{figure}

\section{Convergence Behaviour of Dynamic Fixed Point Mapping}
\label{sec:convergence}
\subsection{Empirical Observations}

Figure \ref{fig:integer_linear} shows that the bit-width, $b$, is controlled by adjusting the number of rounded bits in the rounding function.
Here we study the effect of changing the integer bit-width on the metric performance of the model.

The motivation of varying the bit-width of the dynamic fixed-point is to control the variance induced by the linear fixed-point mapping.
Our experiments show that using dynamic fixed-point with a bit-width of 10 achieves the same performance as the FP32 fine-tuning method.
Figure \ref{fig:ablation} demonstrates the F1 score of fine-tuning BERT on SQuAD v2.0 dataset against the fixed-point bit-width.
Note that the fixed-point arithmetic with a bit-width higher than 10 bits is able to closely match the F1 score of the FP32 baseline, that is indicated by the red line in the figure.
Also note that in our experimental setup for the 8-bit dynamic fixed-point format, we use 12-bit input activations to close the F1 score gap with the FP32 baseline. The reason for using higher bit-width input activations is that we observed 8-bit activation dramatically reduces the F1 score.
Figure \ref{fig:ablation_activation} shows the effect of input activation bit-width on the F1 score when the weights are 8-bit integers.
Changing the bit-width of the input activation from 8 bits to 12 bits significantly increases the F1 score.
Increasing the input activation bit-width beyond 12 bits has a negligible effect on the F1 score, confirming that 12 bits is the minimum required bit-width of the input activations for this application with 8-bit integer weights.

\subsection{Theoretical Analysis}

Here, we study the effect of varying dynamic fixed-point mapping bit-width on the stochastic gradient descent method. The goal is to show the relation of weight and activation bit-widths on the convergence of integer training.
Let us consider the following simplified weight update equation
\begin{equation}
    {w}_{k+1}= {w}_{k} + \bar\eta \hat g(w_k,\xi_k),\
    \label{eq:sgd}
\end{equation}
where $\hat g(w_k,\xi_k)$ is the dynamic fixed-point gradient and $\bar\eta$ is the learning rate during the fine-tuning phase.
Furthermore, we also consider the following common assumptions in sequel.

\noindent\\\textbf{Assumption 1 (Lipschitz-continuity).} The loss function $\mathcal{L}(w)$ is continuously differentiable and its gradients satisfies the following inequality where $L > 0$ is the Lipchitz constant

\begin{align}
    \hspace{-3mm}\mathcal{L}(w) \leqslant \mathcal{L}(\bar w) \nonumber+ &\nabla \mathcal{L}(\bar w)^\top(w-\bar w) \\\nonumber
    &+ \frac{1}{2}L|| w- \bar w||^2_2; \\
    &~~~\forall ~ w,\bar w \in \mathbb{R}^d.
     \label{eq:lip}
     \end{align}

\noindent\\\textbf{Assumption 2.} (i) $\mathcal{L}(w_k)$ is bounded. (ii) $b$-bit dynamic fixed-point gradients $ \hat g(w_k,\xi_k)$ is an unbiased estimator of the true gradients of the loss function $\nabla \mathcal{L}( w_k)^\top \mathbb{E}_{\xi_k}\{ \hat g(w_k,\xi_k) \} = ||\nabla \mathcal{L}(w_k)||^2_2 = ||\mathbb{E}_{\xi_k}\{ \hat g(w_k,\xi_k)  \}||^2_2,$ and (iii)  with the $b$-bit dynamic fixed-point gradients i.e. $\hat{g}(w_k,\xi_k)$, there exist scalars $M \geqslant 0$ , $M_V \geqslant 0$, $M^q \geqslant 0$ and  $M^q_V \geqslant 0$ such that for all iterations of SGD 
\begin{align}
    \nonumber&\mathbb{V}_{\xi_k} \{\hat{g}(w_k,\xi_k)\} \\\nonumber
    &\hspace{5mm}\leqslant  M + M^q + (M_V+ M^q_V )|| \nabla \mathcal{L}( w_k)||^2_2 .
\end{align}
Where $M^q$ and $M^q_V$ denote the added variance of $b$-bit dynamic fixed-point mapping on the true gradient variance.
Also note that in order for \textbf{Assumption 2} (i) to hold true, we use stochastic rounding for back propagation.

Suppose \textbf{Assumption 1} and \textbf{Assumption 2} are true, then inequality \eqref{eq:sgd-moments} follows from \cite[Remark 2]{ghaffari2022integer}
\begin{align}
    \nonumber&\mathbb{E}_{\xi_k}\{\mathcal{L}(w_{k+1})\} - \mathcal{L}( w_k) \\\nonumber   
    &\hspace{5mm}\leqslant-(1-\frac{1}{2} \bar \eta L(M_G+M^q_G)) \bar \eta  ||\nabla \mathcal{L}( w_k)||^2_2\\\nonumber
    & \hspace{40mm}+ \frac{1}{2} \bar \eta^2 L(M+M^q),\\
    & \text{with}~~ M_G := 1+M_V ~~\text{and}~~ M^q_G := 1+M^q_V,
\label{eq:sgd-moments}
\end{align}
which shows the effect of added variance of fixed point mapping, i.e. $M^q_V$ and $M^q$, on each step of the optimizer. 

\noindent\\\textbf{Remark 1.} In inequality \eqref{eq:sgd-moments}, the first term, $-(1-\frac{1}{2} \bar \eta L(M_G+M^q_G)) \bar \eta  ||\nabla \mathcal{L}( w_k)||^2_2$ contribute to decreasing the loss $\mathcal{L}$ while the second term, $\frac{1}{2} \bar \eta^2 L(M+M^q)$,  prevents it. Also note that when $M^q$ and $M^q_G$ are increased, they negatively affect the descent of the loss $\mathcal{L}$. This means for a good convergence behaviour, representation mapping variance bounds, i.e. $M^q$ and $M^q_G$,  must be controlled.

\noindent\\\textbf{Remark 2.} For dynamic fixed-point mapping  with $b$-bit integers, the representation mapping variance bounds i.e. $M^q$ and $M^q_G$, are closely related to the bit-width $b$.
Here, we study these two constants for a linear layer. Let us denote $\hat{\mathbf{A}}$ as the $b$-bit dynamic fixed-point version of tensor $\mathbf{A}$ and $\hat a_{ij}$ as its $ij^{th}$ element. We can relate $\hat a_{ij}$ and $a_{ij}$ with an error term $\delta$ such as $\hat a_{ij} = a_{ij}+ \delta^\mathbf{A}_{ij} $. For a linear layer $\hat{ \mathbf{Y}}=\hat{ \mathbf{X}} \hat{ \mathbf{W}}$, the computation of the $b$-bit dynamic fixed-point gradients in the back propagation is 
\begin{equation}
   \hat{ \mathbf{C}} = \frac{\partial \hat {\mathbf{L}}}{\partial \hat{ \mathbf{W}}} =  \frac{\partial \hat{ \mathbf{Y}}}{\partial \hat {\mathbf{W}}} \frac{\partial \hat {\mathbf{L}}}{\partial \hat {\mathbf{Y}}}
   = \hat {\mathbf{X}}^{\top}\frac{\partial \hat{ \mathbf{L}}}{\partial \hat{ \mathbf{Y}}} = \hat {\mathbf{X}}^{\top} \hat {\mathbf{G}}.
\end{equation}
It is of interest to find the relation between $\hat {\mathbf{C}}=\hat {\mathbf{X}}^{\top} \hat {\mathbf{G}}$ in the integer back propagation and the true gradients $ \mathbf{C}= \mathbf{X}^{\top}  \mathbf{G}$.
We can derive the variance for each element $\hat c_{ij}$ by expanding the error terms $\delta$,

\begin{align}
    \mathbb{V} \{\hat c_{ij}\}&\nonumber
    = \mathbb{V} \left \{ \sum^N_{n=1} \hat x_{ni} \hat g_{nj} \right \} \\\nonumber
    &= \mathbb{V} \left \{ \sum^N_{n=1} (x_{ni} + \delta ^\mathbf{X}_{ni})  (g_{nj} + \delta^\mathbf{G}_{nj}) \right \} \\\nonumber
    &\leqslant \mathbb{V} \left \{ \sum^K_{n=1} x_{ni} g_{nj} \right \} \\\nonumber
    & \hspace{1cm} + \sigma^2_\mathbf{G} \mathbb{E}\{||\mathbf{X}^{\top}_{i.}||^2_2\} + \sigma^2_\mathbf{X} \mathbb{E}\{||\mathbf{G}_{.j}||^2_2\} \\
    & \hspace{1cm} + N\sigma^2_\mathbf{X} \sigma^2_\mathbf{G} \nonumber\\\nonumber
    &= \mathbb{V} \{ {c}_{ij}\} + \sigma^2_\mathbf{G} \mathbb{E}\{||\mathbf{X}^{\top}_{i.}||^2_2\} \\
    & \hspace{1cm} + \sigma^2_X \mathbb{E}\{||\mathbf{G}_{.j}||^2_2\} + N\sigma^2_\mathbf{X} \sigma^2_\mathbf{G}. \nonumber\\
\label{eq:variance}
\end{align}
In inequality \eqref{eq:variance}, $\sigma^2_\mathbf{G} = \mathrm{max}_{i,j} ( \mathbb{V}\{\delta^\mathbf{G}_{i,j}\})$  and $\sigma^2_\mathbf{X} = \mathrm{max}_{i,j} ( \mathbb{V}\{\delta^\mathbf{X}_{i,j}\})$. Also note $||\mathbf{X}^{\top}_{i.}||^2_2= \sum^J_j x_{ji}^2$ denotes the squared L-2 norm of the $i^\text{th}$ \textit{row} of $\mathbf{X}^{\top}$ and $||\mathbf{G}_{.j}||^2_2 = \sum^I_i g_{ij}^2$ denotes the squared L-2 norm of the $j^\text{th}$ \textit{column} of $\mathbf{G}$.
Furthermore, by defining
\begin{equation}
    \begin{cases}
      M^q := \sigma^2_\mathbf{G} (\mathbb{E}\{||\mathbf{X}^{\top}_{i.}||^2_2\} + N\sigma^2_\mathbf{X}) \\
      M^q_{V} :=  \sigma^2_\mathbf{X} \\
    \end{cases}
    \label{eq:constant}
\end{equation}
Equation \eqref{eq:constant} shows that $M^q$ depends on variance of dynamic fixed-point mapping for input activations and gradients while $M^q_G$ only depends on $b$-bit dynamic fixed-point gradients variance.

\begin{table*}[t!]
\centering
\begin{tabular}{c|c|c|c|c|c|c|c|c|c}
                                                              & QQP  & QNLI  & MNLI  & SST-2 & STSB & RTE & MRPC & CoLA & Average  \\\hline
\textbf{FP32} & 91.0/88.0 & 91.1  & 84.2 & 92.5 & 88.3 & 63.8  & 82.5/87.8 & 57.2 & 82.6 \\ \hdashline 
\textbf{FP16 AMP} & 90.9/87.9  & 91.2  & 84.1 & 92.4 & 88.3 & 64 & 82.1/87.7 & 57.5 & 82.6 \\ 
\textbf{16-bit integer} & \textbf{91.0/88.0} & 91.2  & \textbf{84.2} & 92.5 & \textbf{88.3} & \textbf{64.5} & \textbf{82.3/87.6} & \textbf{57.7} & \textbf{82.7} \\
\textbf{12-bit integer} & 90.9/88.0 & \textbf{91.2} & 84.0  & \textbf{92.6} & 87.9 & 63.5 & 81.3/87.4 & 56.7 & 82.4 \\
\textbf{10-bit integer} & 90.8/87.8 & 91.0 & 84.0 & 92.5 & 87.5  & 62.7 & 78.4/85.8  & 57.6 & 81.8  \\
\textbf{8-bit integer} & 90.1/86.8  & 90.8  & 83.7 & 92.3 & 87  & 61.8 & 76.8/84.7  & 55.0 & 80.9  \\
\end{tabular}
\caption{Metric performance of integer fine-tuning of BERT on selected GLUE tasks. The reported metric for QQP and MRPC is accuracy and F1 score, for QNLI, MNLI, RTE, and SST-2 is accuracy, for STSB is the Pearson-Spearman correlation, and for CoLA is the Matthews correlation.}
\label{tab:glue}
\end{table*}

\noindent\\\textbf{Proposition 1.} For dynamic fixed-point representation of tensor $\hat {\mathbf{A}}$ with $b$-bit integers, the variance of error for element $i$ satisfies the following inequality

\begin{equation}
      \mathbb{V}\{\delta_i^\mathbf{A}\} \leqslant 2^{2(e_\mathrm{scale_\mathbf{A}}-b+2)}.
    \label{eq:var-bound}
\end{equation}
\noindent\textbf{\textit{Proof}.} Using dynamic fixed-point mapping to $b$-bit integers, the error $\delta_i^\mathbf{A}$ satisfies the following bound
\begin{equation}
\begin{gathered}
    - 2^{e_\mathrm{scale_\mathbf{A}}} \underbrace{(0.000001)_2}_{b-1}\leqslant \delta_i^\mathbf{A} \leqslant 2^{e_\mathrm{scale_\mathbf{A}}}\underbrace{(0.000001)_2}_{b-1}\\
    - 2^{e_\mathrm{scale_\mathbf{A}}-b+2}\leqslant \delta_i^\mathbf{A} \leqslant 2^{e_\mathrm{scale_\mathbf{A}}-b+2}.      
\end{gathered}
    \label{eq:delta-bound}
\end{equation}
Thus, the inequality \eqref{eq:var-bound} is obtained by using Popoviciu's inequality on variances
\begin{align}
    \nonumber\mathbb{V}\{\delta_i^\mathbf{A}\} & \leqslant \frac{1}{4} (2^{e_\mathrm{scale_\mathbf{A}}-b+2} - (-2^{e_\mathrm{scale_\mathbf{A}}-b+2}))^2\\ 
    &\leqslant 2^{2(e_\mathrm{scale_\mathbf{A}}-b+2)}.     
    \label{eq:var-bound-proof}
\end{align}

\noindent\textbf{Remark 3.} Inequality \eqref{eq:var-bound} shows that increasing bit-width $b$ in dynamic fixed-point mapping reduces the variance of the error. This confirms our experimental results on SQuAD v2.0 dataset that for $b>10$, F1 score can match FP32 baseline, see Figure \ref{fig:ablation}. Also note in equation \eqref{eq:constant}, both $M^q$ and $M^q_V$ depend on $b$-bit dynamic fixed-point mapping variance of input activation $\sigma^2_\mathbf{X}$. Hence, increasing $b$ for input activations while keeping weights in 8-bit format must improve the convergence behaviour. This phenomenon is also confirmed by our experimental results on SQuAD v2.0 dataset demonstrated in Figure \ref{fig:ablation_activation}.

\begin{table}[t]
\centering
\begin{tabular}{c|c|c}
                    & SQuAD v1.1 & SQuAD v2 \\ \hline
\textbf{FP32} & 80.5/88.0 & 70.6/73.8 \\ \hdashline
\textbf{FP16 AMP} & 79.9/87.6 & 70.6/73.9 \\
\textbf{16-bit integer} & \textbf{80.7/88.0} & \textbf{70.6/73.9} \\
\textbf{12-bit integer} & 79.8/87.6 & 70.5/73.8 \\
\textbf{10-bit integer} & 78.4/86.6 & 69.8/73.2 \\
\textbf{8-bit integer}  & 75.6/84.5 & 65.5/69.2 \\
\end{tabular}
\caption{Metric performance of fine-tuning BERT on SQuAD v1.1 and v2.0 datasets. For both datasets the exact match metrics and F1 scores are reported.}
\label{tab:squad}
\end{table}

\section{Experimental Results} 
\label{sec:results}

\subsection{Experimental Setup}

We fine-tuned BERT base on a series of downstream tasks to compare the performance of our integer fine-tuning method with FP16 and FP32 fine-tuning methods.
FP16 AMP setup uses NVIDIA's automatic mixed precision\footnote{https://developer.nvidia.com/automatic-mixed-precision} and the FP32 baseline is the default implementation from Pytorch.

The model is fine-tuned on selected tasks of GLUE benchmark \cite{wang2018glue}, along with the Stanford Question Answering Datasets, i.e. SQuAD v1.1 and SQuAD v2.0 \cite{rajpurkar2016squad}.

All the fine-tuning setups use the same hyper-parameters and are fine-tuned for the same number of epochs.
Each reported metric is the average of five runs with five different random seeds to mitigate the effects of random variation of the results.
The fine-tuning experiments are performed based on the fine-tuning scripts of the Hugging Face library \cite{wolf2019huggingface}.
For GLUE experiments the fine-tuning is performed for 5 epochs and the learning rate is set to $2 \times 10^{-5}$. Also, the per-device fine-tuning batch-size is set to 32. Fine-tuning BERT on SQuAD datasets is done for 2 epochs and the learning rate is $5 \times 10^{-5}$ and the per-device fine-tuning batch-size is 12. All experiments are run on eight NVIDIA V100 GPUs with 32 gigabytes of VRAM.

\subsection{Results}

The results of fine-tuning BERT base on GLUE benchmark and SQuAD datasets are presented in Table \ref{tab:glue} and Table \ref{tab:squad} respectively.
GLUE benchmark contains a series of downstream tasks, designed to evaluate a diverse set of language understanding abilities of NLP models.
SQuAD datasets contain a series of text passages accompanied by a question and the task is to predict the span of the answer in the passage.
Using 16-bit integer data type, BERT is able to either match or outperform the FP32 performance for all tasks.
The 16-bit integer BERT also shows similar or better performance compared to the FP16 mixed precision fine-tuning method.
Further reducing the integer bit-width to 8, fine-tuning BERT exhibits an average of 1.7 point drop on GLUE benchmark and 4.5 point drop for SQuAD datasets.
Moreover, our experiments show that using 10-bit and 12-bit integers has average score drops of 0.8 and 0.3 points for GLUE tasks, and 0.8 and 0.2 point for SQuAD datasets respectively.

\begin{figure}[t]
\centering
\includegraphics[width=0.9\columnwidth]{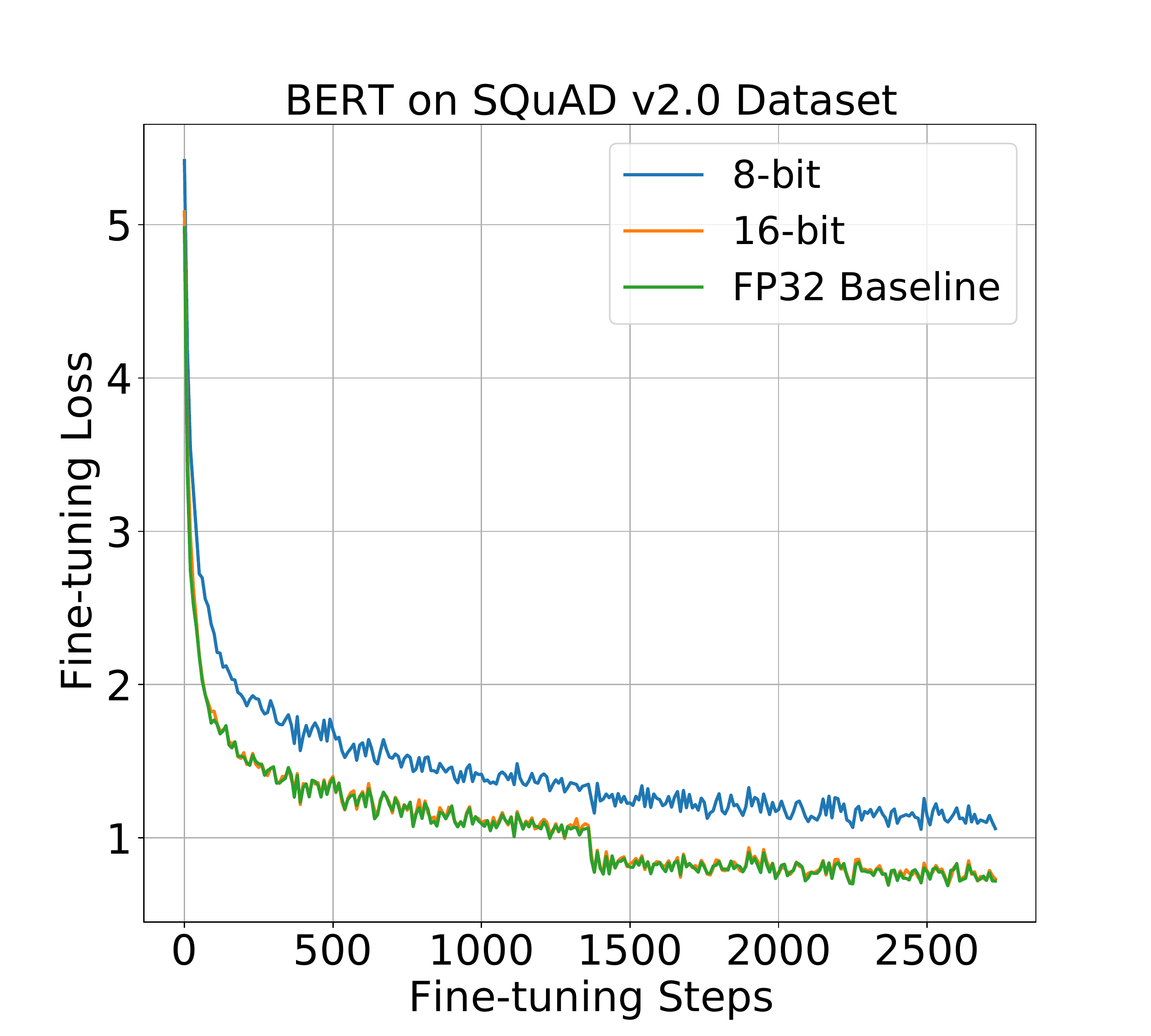}
\caption{Integer fine-tuning loss trajectory of BERT on SQuAD v2.0 dataset for 2750 iterations.}
\label{fig:loss}
\end{figure}

\subsection{Loss Trajectory}
Figure \ref{fig:loss} shows the loss trajectory of integer fine-tuning BERT on SQuAD v2.0 dataset using 16-bit and 8-bit integers, along with FP32 method.
The fine-tuning loss trajectory of BERT using 16-bit integer closely follows  the FP32 loss trajectory.
On the other hand, when fine-tuning with 8-bit integer parameters and 12-bit integer input activations, the loss trajectory is slightly shifted, but follows the same trend of its FP32 counterpart.

\section{Conclusion}
\label{sec:conclusion}

We proposed an integer fine-tuning method for transformer based language models using dynamic fixed-point format.
We used dynamic fixed-point data type to represent parameters, input activations and gradients in integer values.
As a result, our fine-tuning method uses integer arithmetic for the forward and back propagation of compute intensive layers such as linear, layer-norm and embedding layers of BERT model.
Furthermore, we studied that increasing the bit-width of the dynamic fixed-point format reduces the variance of the mapping function and thus, improves the convergence of our integer fine-tuning method.
We conduct fine-tuning experiments on GLUE benchmark and SQuAD datasets to compare the metric performance of our integer BERT with FP16 mixed precision and FP32 fine-tuning methods.
Our experiments show that the 16-bit integer fine-tuning is able to achieve the same metric performance as the FP16 mixed precision fine-tuning method.
In addition, fine-tuning BERT with lower bit-width data types, i.e. 8-bit integer, maintains an average drop of metric score within 3.1 points of the FP16 setup. 

\section*{Limitations}

Although our integer fine-tuning method uses integer numbers for compute intensive layers of BERT, integer support for non-linear layers of BERT, e.g. softmax and GELU activation, are left for future work.


We have shown in Figure \ref{fig:energy} that the integer data types are faster for the general case.
However, a direct comparison of the time and memory cost of our integer fine-tuning method with the FP16 and FP32 methods is left for future works due to lack of access to a proper hardware with integer tensor core support.

Despite the similarities between fine-tuning and pre-training phases, they differ in key aspects of training such as dataset size and number of epochs.
The challenges of using integer arithmetic in the pre-training phase will be studied in the future work.


\bibliography{anthology,custom}
\bibliographystyle{acl_natbib}

\appendix

\section{Data Types}
\label{sec:appendix_datatypes}

In this section we provide a brief overview of various data types mentioned in this work.

Floating-point data type is used to represent decimal fractional numbers.
A binary floating-point number has three components of sign ($s$), mantissa ($m$), and exponent ($e$).
Using these components, floating-point number $x$ is shown as:
\begin{equation*}
    x = (-1)^s \times m \times 2^{e - t}
\end{equation*}
where $t$ is the precision and $0 \leq m \leq 2^t - 1$.
Another way of representing floating-point numbers is as
\begin{equation*}
    x = (-1)^s \times 2^e (\frac{d_1}{2} + \frac{d_1}{4} + \ldots + \frac{d_t}{2^t})
\end{equation*}
where $d_i$ are binary digits of m.
For FP32, exponent and mantissa are 8 and 23 bit integer numbers.

Fixed-point is another data type for representing fractional numbers.
Unlike floating-point numbers where each mantissa is scaled using its respective exponent, fixed-point uses a single scale factor for all the numbers.

We use the dynamic fixed-point data type in our integer fine-tuning method.
Also known as block floating-point, this format uses a different scale for each block of numbers to allow for more flexibility.

\end{document}